# Comparison of Deep Learning Segmentation and Multigrader-annotated Mandibular Canals of Multicenter CBCT scans

DDS Jorma Järnstedt[1], MSc (Tech.) Jaakko Sahlsten[2], MSc (Tech.) Joel Jaskari[2], DPhil Kimmo Kaski[*,2,5], DDS Helena Mehtonen[1], MSc Ziyuan Lin[2,3], PhD Ari Hietanen[3], MSc (Econ.) Osku Sundqvist[3], MSc (Tech.) Vesa Varjonen[3], MSc (Tech.) Vesa Mattila[3], DDS, MSc Sangsom Prapayasotok[4] & DDS, MSc, PhD Sakarat Nalampang[4]

[1]Medical Imaging Centre, Department of Radiology Tampere University Hospital, Teiskontie 35, 33520 Tampere, Finland
[2]Aalto University School of Science, Maarintie 8, 02150 Aalto, Finland
[3]Planmeca Oy, Asentajankatu 6, 00880 Helsinki, Finland
[4]Division of Oral and Maxillofacial Radiology, Faculty of Dentistry, Chiang Mai University, Suthep Rd., T. Suthep, A. Muang, Chiang Mai, Thailand
[5]Alan Turing Institute, British Library, 96 Euston Rd, London NW1 2DB, UK

*corresponding author

# Abstract

Deep learning approach has been demonstrated to automatically segment the bilateral mandibular canals from CBCT scans, yet systematic studies of its clinical and technical validation are scarce. To validate the mandibular canal localization accuracy of a deep learning system (DLS) we trained it with 982 CBCT scans and evaluated using 150 scans of five scanners from clinical workflow patients of European and Southeast Asian Institutes, annotated by four radiologists. The interobserver variability was compared to the variability between the DLS and the radiologists. In addition, the generalization of DLS to CBCT scans from scanners not used in the training data was examined to evaluate the out-of-distribution generalization capability. The DLS had lower variability to the radiologists than the interobserver variability between them and it was able to generalize to three new devices. For the radiologists' consensus segmentation, used as gold standard, the DLS had a symmetric mean curve distance of 0.39 mm compared to those of the individual radiologists with 0.62 mm, 0.55 mm, 0.47 mm, and 0.42 mm. The DLS showed comparable or slightly better performance in the segmentation of the mandibular canal with the radiologists and generalization capability to new scanners.

**Keywords**

# Introduction

Recently there has been a rapid increase of studies demonstrating that Artificial Intelligence methodologies, especially those based on deep learning neural networks, can distinguish structural patterns in medical imaging data with excellent accuracy[1], and serve radiologists as augmenting tools for clinical workflow[2]. However, in dental and maxillofacial radiology, so far there are relatively few studies where deep learning approaches have been used for localizing or segmenting the bilateral mandibular canals[3–8], which host the inferior alveolar nerves, each containing an artery and a vein. These canals have two openings; a foramen mandibulae posterior in the ramus area, and foramen mentale anterior in the parasympheal area. What makes the canal localization in CBCT images challenging is that there are a number of anatomical variations in the pathway and shape of the canal, and also ethnic variability is known to play a role[9,10]. As the inferior alveolar nerves supply motor and sensory innervations, any damage to them can cause temporary or permanent nerve injuries. Therefore, the accurate knowledge of mandibular canal locations is extremely important for various oral and maxillofacial surgical operations, and in the diagnosis of neurogenic, vascular, or adjacent lesions to the canals.

The first studies on using deep learning convolutional neural network (CNN) approach to automatically localize the bilateral mandibular canals in CBCT images appeared as recently as 2020[3,4]. In the study by Kwak et al.[3] a 2D SegNet and 2D as well as 3D U-Net CNNs were used on images of 102 patients from a single scanner and the performance was measured using the mean intersection over union with a value of 0.577. At the same time Jaskari et al.[4] applied a 3D U-net style CNN approach to a large number of clinically heterogeneous CBCT images of 637 patients using four different scanners and as performance measures they reported the average symmetric surface distance value of 0.45 mm, mean curve distance value of 0.56 mm, and Dice similarity coefficient (DSC) value of 0.61. The authors demonstrated that their method can produce excellent localization accuracy of the mandibular canals with good robustness and concluded that integrating this approach into clinical workflow could significantly reduce the load of radiologists' manual labor involved in the mandibular canal annotations. Recently, CNNs have also been used with a smaller number of patient images but from multiple CBCT scanners. As a matter of fact, Lahoud et al.[5] used a CNN with 235 patients images from four scanners and Lim et al.[6] with 138 patient images from three scanners. Lahoud et al.[5] reported the DSC value of 0.774 and Lim et al. [6]the DSC of 0.580 and also the mean intersection over union value of 0.636. In addition, there are further two deep learning based studies also using smaller numbers of patient images from a single CBCT scanner, i.e. a study by Dhar et al.[7] with 187 patients reporting the Dice similarity coefficient value of 0.57, the mean intersection over union value of 0.70 and the mean curve distance value of 0.62 mm as the performance measures. In a study by Wei et al.[8] a CNN was used with 81 patients using multiplanar reconstruction images and reported the DSC value of 0.93, average symmetric surface distance value of 0.16 mm and the mean curve distance value of 1.59 mm as the performance measures. Although the previous works have shown promising results, the datasets have had limited clinical diversity based on ethnicity and patient age groups, and there have been only limited or no description on patient specific heterogeneities and the number of CBCT scanners. In the literature, there are a variety of generally used and valid performance measures such as the Dice similarity coefficient, the mean intersection over union, the average symmetric surface distance, and the mean curve distance. Of these the first three measure the segmentation performance and the fourth measures the deviation between the paths of two curves, namely between the original ground truth and the prediction curves[4,7,8].

In the present study, we focus on validating a deep fully convolutional neural network-based mandibular canal segmentation system, introduced in[4], from the clinical and technical points of view. To comprehensively analyze the system, we use a larger and clinically more heterogeneous dataset of CBCT scans than reported in any of the previous studies. Indeed, our database has images from five different CBCT scanners from four different vendors and patient cohorts of two different ethnicities that we use to train a new version of the deep learning system. In the clinical validation process, we compare the performance of the deep learning system against four experienced dental and maxillofacial radiologists. In addition, we estimate the interobserver variability between the radiologists, to evaluate the differences in the canal localization by radiologists, and how it compares to the variability between the deep learning system and the radiologists. For technical validation, we analyze the out-of-distribution generalization capability of a version of the deep learning system[4], which was trained with 509 images from Planmeca ProMax and Scanora 3Dx devices. This includes temporal and geographical generalization, as we analyze three different devices with a portion of the imaged patients from another country as well as more recent images from previously used devices. To evaluate the quantitative performance between the deep learning system and the radiologists, we construct consensus segmentations from the four radiologists' segmentations, and evaluate it and the radiologists on them. Moreover, to analyze the qualitative performance, a senior radiologist evaluated the automatic and expert segmentations to identify the types of errors in them. To the best of our knowledge, this is the first time in dental and maxillofacial radiology CBCT imaging, when the out-of-distribution generalization capability and performance of a deep learning neural network approach is compared to interobserver variability.

# Methods

## Data Collection and Cohort Description

The CBCT imaging data was acquired from the Cranio and Dentomaxillofacial Radiological Department of the University Hospital of Tampere (TAUH), Finland as the first cohort, and from the Department of Oral Radiology, Faculty of Dentistry, Chiang Mai University (CMU), Thailand as the second cohort. All the data in this study is from a normal clinical workflow that represents pre- and postoperative examinations of patients of 10 to 95 years old. The reasons for radiological examination include normal findings and anatomy, but also various traumas, benign or malign pathological conditions, and syndromes. The CBCT scans were randomly and retrospectively selected and pseudonymised, before the annotation process.

The collected dataset of CBCT scans consisted of 1103 individuals, with 869 Finnish patients (79%) and 234 Thai patients (21%). In the Finnish population, with the mean age of 53.7 years, 56% were females and 44% males. The Thai patient population with the mean age of 39.8 years, consists of 51% females and 49% males. In the dataset, 649 CBCT scans (57%) were imaged using Planmeca ProMax 3D Max/Mid, 125 (11%) using Planmeca Viso G7, 124 (11%) using Soredex Scanora 3Dx, 120 (11%) using NSTDA DentiScan, and 114 (10%) using NewTom GiANO HR. The first three scanners were used in the first cohort and the latter two scanners in the second cohort. The scan resolution ranged from 0.1 to 0.6 mm isotropic voxel-sizes with most commonly 0.2 mm (59%), 0.4 mm (23%) or 0.3mm (13%). All volumes were resampled into 0.4 mm isotropic voxel spacing using linear interpolation before deep learning analysis. Human annotators had access to the scans in the original resolution and could augment the view of the scans using software tools.

The deep learning methodology is driven by data such that in order to estimate the performance of the model, a partition of the data is held out and only used for validation of the results. This set is called the test set and it was randomly selected with uniform distribution of scanners. The rest of the data was used for the deep learning system development, and thus called the development set. Flowchart for data collection is presented in Figure 1.

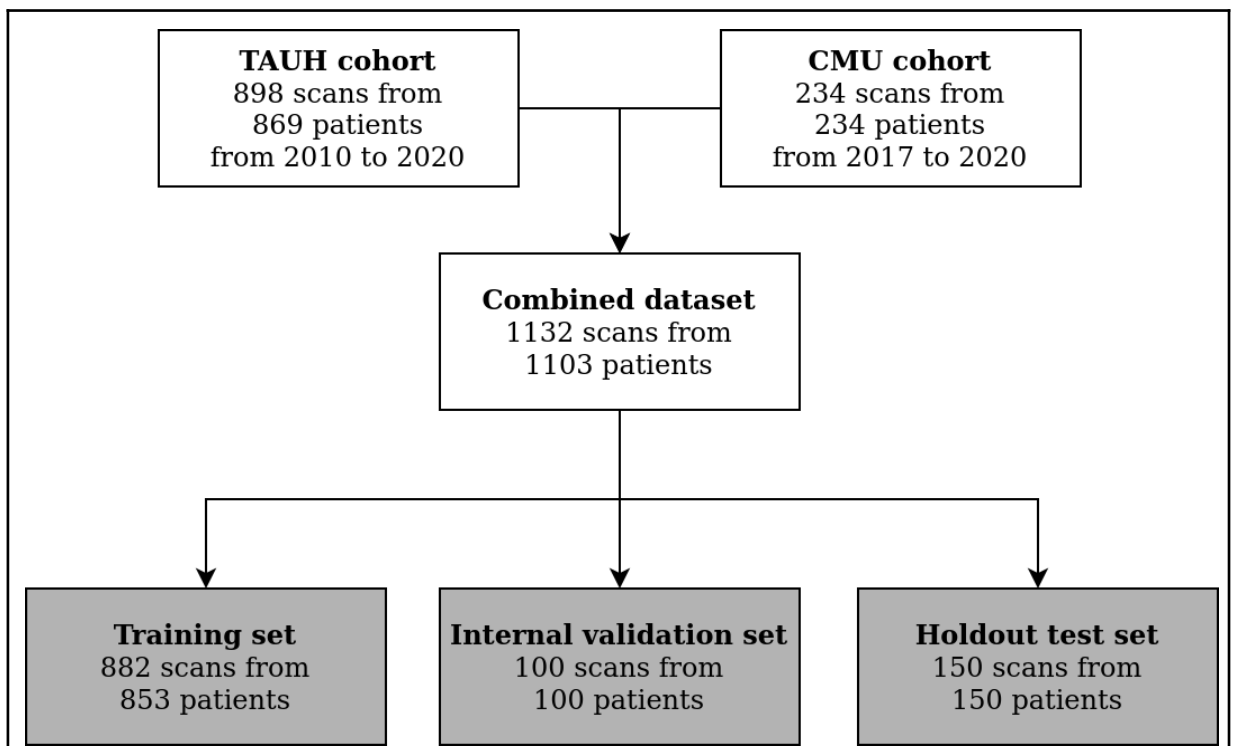

**Figure 1**: Flowchart for data collection. Patients were selected from recent clinical workflows at the Cranio and Dentomaxillofacial Radiological Department of the University

Hospital of Tampere, Finland (TAUH), and from the Department of Oral Radiology, Faculty of Dentistry, Chiang Mai University (CMU), Thailand. Combined dataset had a total number of scans 1132 from 1103 patients that were splitted into training set, internal validation set and holdout test set with 853, 100 and 150 patients, respectively.

The mandibular canal was annotated using Planmeca developed Romexis® 4.6.2. software, which has a built-in tool for mandibular canal annotation using control points and spline interpolation. The control points were standardized to be 3 mm apart from each other, and in foramen mentale curvature area the canal was annotated using multiple control points, when necessary. The annotation of the development set was performed by two dentomaxillofacial radiologists referred to as Expert3 and Expert4, one with over 35 and 11 years of experience in dentistry with the Expert3 annotating approximately 90% of the set and Expert4 the rest. The test set was annotated, in addition to the previous annotators, by two other dentomaxillofacial radiologists referred to as Expert1 and Expert2, one with over 35 and another with 14 years of experience in dentistry. The test set was annotated independently by all the radiologists. Since the end-point of the mandibular canal in the mandibular foramen region is ambiguous, the end-point was selected to match the shortest annotation in the superior direction, thus improving the sensitivity of the canal localization in the foramen mentale and dental regions rather than highlighting differences in canal path lengths. The set was also subjectively annotated for the clarity of mandibular canal visibility to be either Clear or Unclear. The test set scans were also annotated for the following conditions if present: movement artifact, bisagittal osteotomy, metal artifact, difficult pathology and difficult bone structure including difficult anatomy and osteoporosis.

## Validation of the Deep Learning System

We utilize the previously proposed CNN-based method[4] as our deep learning model, i.e. a type of fully convolutional neural network using a U-net style architecture[11]. The model utilizes three dimensional convolutional layers that enables the recognition of patterns in axial, coronal, and sagittal planes, simultaneously. We develop an improvement to the canal extraction post processing algorithm from the model segmentation. In short, mandibular canal route segments are obtained from the CNN output using a skeletonization routine[12], route segments are concatenated using a heuristic, and then the routes with anatomical characteristics of mandibular canals are selected. Finally, a pair of routes with most symmetricity is chosen as the pair of canals.

## Statistical Analysis

For the evaluation of localization performance, we used the mean curve distance (MCD) measure, similar to previous works[4,7,8], and in addition we propose the symmetric mean curve distance (SMCD) measure to summarize the bi-directional differences in two estimates of the canal. In the case of MCD, for each point on a *ground truth* curve, the distance to the closest point on another *estimator* curve is computed and then these distances are averaged. Thus, it estimates the average deviation between the paths of two curves from the point of view of the *ground truth* curve, effectively measuring the *sensitivity* in the curve localization. As the MCD is unidirectional, it does not capture changes in precision, e.g. when the *estimator* curve is longer than the *ground truth* curve. Hence, we propose the SMCD measure, which is calculated as the average of the MCD values computed both ways and effectively accounts for both the *precision* and *sensitivity* in the curve localization. The SMCD is also useful in summarizing the interobserver variability, as the role of the *ground truth* and *estimator* curves are not well defined.

In mathematical terms, let *T* and *E* be the sets of points that define the *ground truth* curve and the *estimator* curve, respectively. The *point to curve distance* function *d(x,S)* is defined such that it computes the minimum Euclidean distance from a point *x* to the set of points *S* that defines a curve:

$$d(x, S) = \min_{s \in S} \|x - s\|_2 . \tag{1}$$

Then the MCD is computed as:

$$MCD(T, E) = \frac{1}{|T|} \sum_{t \in T} d(t, E) . \tag{2}$$

The SMCD is computed using Equation (2) and a permutation of the arguments:

$$SMCD(T, E) = \frac{1}{2}(MCD(T, E) + MCD(E, T)) \,. \qquad (3)$$

We have also evaluated the proportion of the canal path within a 2 mm safety margin that is used as a guideline for implant planning[13]. Volumetric segmentation measures, such as the Dice similarity coefficient, were deemed unsuitable for our main results, since the annotation tool in our study was designed to use a fixed diameter for the canal. However, the results measured with the Dice similarity coefficient are reported in the Supplementary Figures S2-S5.

We evaluated and analyzed the variability between experts and the deep learning model by comparing the radiologist canal annotations and the segmentations produced by the deep learning system in a pairwise manner. In order to estimate the highest level of interobserver variability, for each CBCT scan we selected the pair of expert annotations with the highest mean curve distance. Similarly, we selected one expert annotation with the highest mean curve distance with the deep learning system, by treating the expert annotation as the ground truth curve and the automatic segmentation as the estimator, to estimate the highest variability between the system and a human expert. The generalization capacity of the previously published system[4] was evaluated similarly by selecting the expert annotations with the highest mean curve distance to the segmentation produced by the system.

We estimated the objective performance between the experts and the deep learning model by constructing a label voting scheme of the expert annotations as the reference 'ground truth' using SimpleITK library[14] and evaluated the performance using SMCD. Specifically, voxels are assigned a background or mandibular canal label with maximum votes and undecided voxels are marked as a canal label. After this the segmentation was skeletonized and the curve was determined with connected component analysis. Statistical significance of all the main results were computed using the Wilcoxon signed-rank test using Python package[15].

## Inclusion/Exclusion Criteria

There was no exclusion criteria in the diagnostically acceptable datasets.

# Results

## Patient Cohorts

The dataset used in this study includes 982 and 150 CBCT scans from 953 and 150 patients for the development and test set, respectively. All reported results are computed for the test set, in which each scan included two mandibular canals. The sample characteristics are presented in Table 1 while the distribution of the devices of the subsets are shown in Table 2 for the development set and holdout test set. The deep learning system training and internal validation sets are described in detail in Table S1.

**Table 1**: Characteristics of the study sample with percentages in parentheses[a].

| Parameter | Cohort 1 (n = 869) | Cohort 2 (n = 234) |
|---|---|---|
| Age (y)[b] | 54.9 ± 33.2 | 39.8 ± 17.0 |
| Gender | | |
|   Female | 487 (56) | 119 (51) |
|   Male | 382 (44) | 115 (49) |
| Race | | |
|   Caucasian | 869 (100) | 2 (1) |
|   Asian | 0 (0) | 232 (99) |

[a]Cohort 1 from University Hospital of Tampere (TAUH) and Cohort 2 from Chiang Mai University (CMU).

[b]Mean and standard deviation.

Table 2: Distribution of scans with five CBCT-scanners and patients for each of the subsets.

| Manufacturer and device | Development | | Holdout | |
|---|---|---|---|---|
| | Scans | Patients | Scans | Patients |
| Planmeca ProMax (3D, 3D Max, 3D Mid) | 619 (63%) | 590 (62%) | 30 (20%) | 30 (20%) |
| Planmeca Viso G7 | 95 (10%) | 95 (10%) | 30 (20%) | 30 (20%) |
| Soredex Scanora 3Dx | 94 (10%) | 94 (10%) | 30 (20%) | 30 (20%) |
| NSTDA DentiScan | 90 (9%) | 90 (9%) | 30 (20%) | 30 (20%) |
| NewTom GiANO HR | 84 (9%) | 84 (9%) | 30 (20%) | 30 (20%) |
| **Total** | **982** | **953** | **150** | **150** |

Unambiguous heterogeneities were marked with at least one expert label: movement artifact (7, 2%), metal artifact (9, 3%) and bisagittal osteotomy (4, 1%). More ambiguous heterogeneities were assessed by using majority voting: difficult bone structure including difficult anatomy and osteoporosis (99, 33%), difficult pathology (3, 1%). There were 275 (92%) canals marked Clear and 25 (8%) Unclear, also determined using majority voting. Comparison of performance in each heterogeneity group is reported in Supplementary Figure S1.

## Interobserver variability

The results of the individual variability analysis are presented in Figure 2, where the experts and the deep learning system are compared in a pairwise manner for all the possible combinations, to obtain a fine-grained analysis. In Figure 2a, the MCD results are presented. The deep learning system and the Expert4 had overall the highest agreement measure with 0.46 mm median MCD and the lowest agreement with the Expert1 with 0.69 mm median MCD. In terms of the interobserver variability, the highest agreement was with the Expert3 and Export4 with 0.48 mm median MCD and the lowest agreement between the Expert1 and Expert3 with 0.70 mm median MCD. The proportion of the canal within the specified 2 mm safety margin, as evaluated in a pairwise manner, is presented in Figure 2b. On average, all the observers had very similar performance, indicating that most of the time the annotated canal paths did not vary more than 2 mm, as the largest difference between Expert1 and Expert2 was 96%.

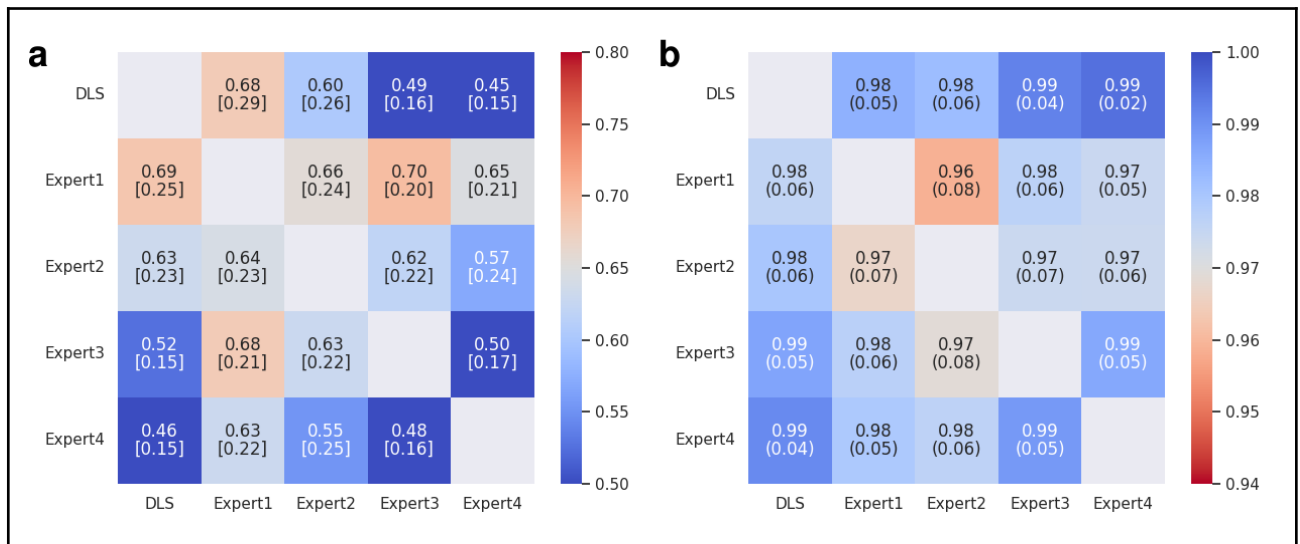

**Figure 2**: Pairwise comparison of the experts and the deep learning system (DLS). Each row represents which assessment was used as the ground truth and each column which was used as the estimate. The mean curve distance measure is asymmetric, which results in asymmetric matrices. (**a**) Median and interquartile range [IQR] of curve distances in millimeters. (**b**) The mean (SD) of the proportion of the canal within a 2.0 mm safety margin.

In the evaluation of the highest variability, there was a statistically significant difference in the mean rank of highest interobserver and model variability ($P<0.001$) with the deep learning system having lower median SMCD. The median [interquartile range] and mean (standard deviation) of highest expert to expert, i.e. interobserver, SMCD were 0.77 [0.25] mm and 0.84 (0.28) mm, respectively, whereas the highest deep learning system to expert SMCD median [interquartile range] was 0.74 [0.28] mm and mean (standard deviation) 0.81 (0.41) mm. In device-wise comparison, the DLS had a non-significant difference in highest variability performance against experts with Scanora3Dx ($P=0.84$) and Viso ($P=0.10$), a significant difference with lower median SMCD on DentiScan ($P<0.001$) and GiANO ($P<0.001$) and ProMax ($P<0.001$). Full results are presented in Figure 3 with scatterplot comparison presented in Supplementary Figure S4.

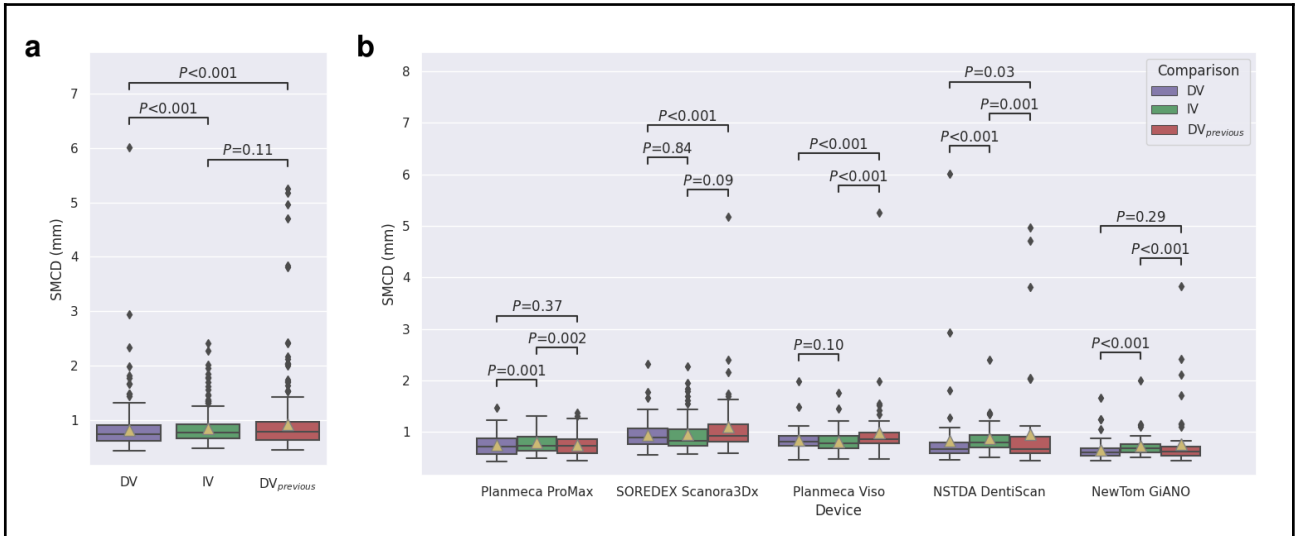

**Figure 3**: Tukey's boxplot comparison of interobserver variability (IV), DLS to expert variability (DV), and previous method[4] to expert variability (DV$_{previous}$), measured in symmetric mean curve distance (mm). Statistical significance measured with Wilcoxon signed-rank test. (**a**) Comparison of full test dataset (N=300). (**b**) Device-wise comparison between the groups (N=60 per device).

The previously published system was able to produce results for 299 out of 300 canals with the highest variability analysis having a median of SMCD of 0.78 [0.33] mm and mean (standard deviation) 0.91 (0.63) mm, with a significant difference to the current system ($P<0.001$). In device-wise comparison, the previously reported system had a non-significant difference in highest variability performance against experts with Scanora3Dx ($P=0.09$), a significant difference with lower median SMCD on DentiScan ($P<0.001$) and GiANO ($P<0.001$) and higher median SMCD on ProMax (P=0.002) and Viso ($P<0.001$). Full results are presented in Figure 3.

## Reference segmentation

There was a statistically significant difference in performance between all the experts and the deep learning system against the reference segmentation ($P<0.001$). Expert1 had the largest median [interquartile range] of SMCD with 0.62 [0.23], Expert2 with 0.55 [0.22], Expert3 with 0.47 [0.14], and Expert4 the lowest with 0.42 [0.14] mm, whereas the deep learning system had the smallest median SMCD of 0.39 [0.11] mm. In addition, DLS had lowest mean (standard deviation) of SMCD with 0.46 (0.39) mm, while the Expert1 had mean of SMCD of 0.68 (0.38), Expert2 0.62 (0.39), Expert3 0.52 (0.38), and Expert4 0.47 (0.40) mm. There were significant missing parts in the reference segmentation for a total of two canals, as there was no agreement between the expert segmentations. These are seen as two outliers with highest SMCD results for all the experts and amongst the outliers of the DLS. The results are presented in Figure 4a and Table S2.

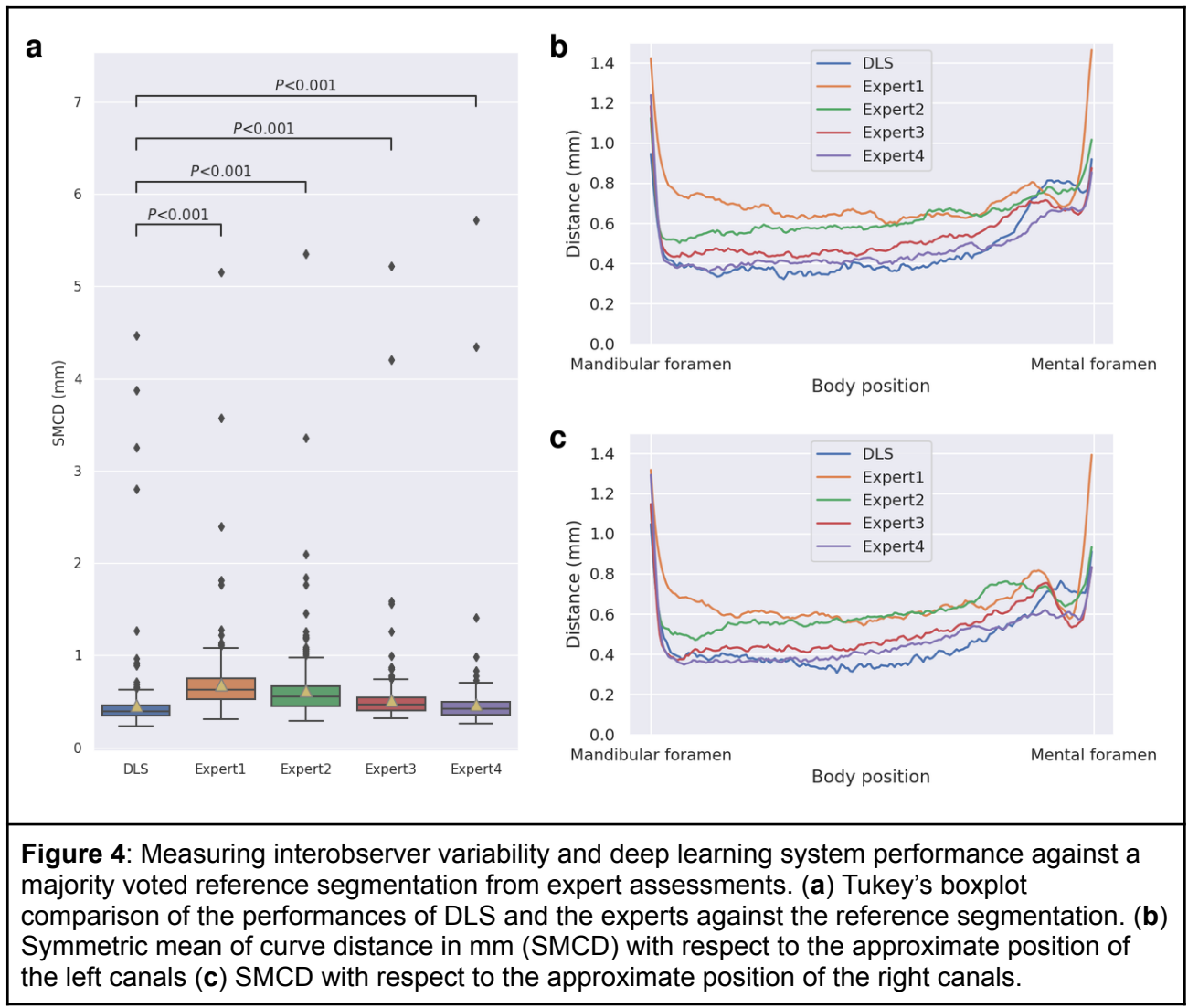

**Figure 4**: Measuring interobserver variability and deep learning system performance against a majority voted reference segmentation from expert assessments. (**a**) Tukey's boxplot comparison of the performances of DLS and the experts against the reference segmentation. (**b**) Symmetric mean of curve distance in mm (SMCD) with respect to the approximate position of the left canals (**c**) SMCD with respect to the approximate position of the right canals.

We also examined, for both the left and right canals, how the distance to the reference annotation is related to on which approximated position of the mandibular canal curve it is measured. In practice, 200 uniformly spaced interpolation points were used to have a dense representation of the curve, and the point to curve distance was computed using Equation (1). The deep learning system performance is similar to the expert performance for both the left and right canals such that the deep learning system was found to perform better than the experts near the mandibular foramen for right canals. There are no major differences in the deep learning system and the expert performances between the left and the right canals. We found that near the foramina, the standard deviation of the curve distance is significantly larger for the deep learning system due to few erroneous cases in the Unclear subset. Visual illustration of the average performance is presented in Figure 4b for the left canal and in Figure 4c for the right canal.

## Qualitative assessment

The qualitative analysis of the test images was done visually by a senior radiologist, who compared the deep learning outcome and the annotations of each radiologist. The deep learning system produced three significant errors (>1/3 of the canal was missing) out of 300 evaluated with one segmentation error and two post-processing errors. The likely cause for the segmentation error was a technical artifact of the CBCT machine. Other 297 out of 300 (99%) canals analyzed were correct. Example segmentation comparing the deep learning system and experts are presented in Figure 5.

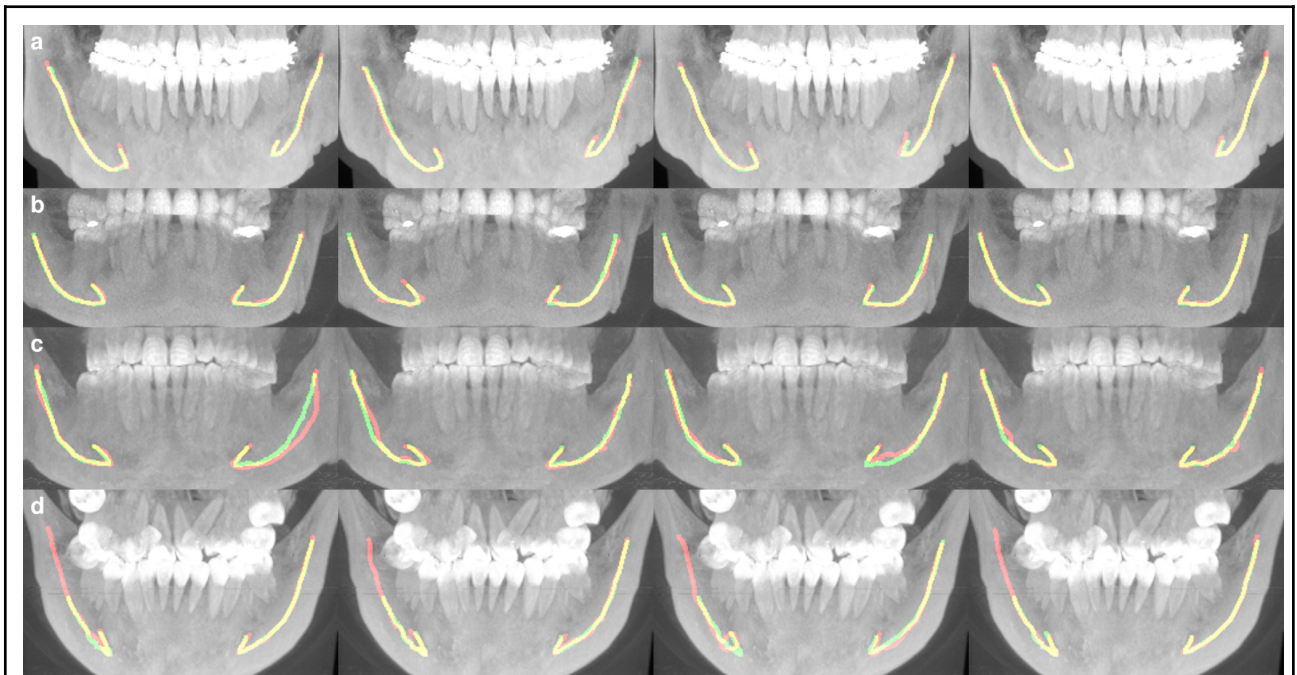

**Figure 5**: Illustration of expert and DLS segmentations. Every image on each column is annotated by the same expert, shown in red, DLS annotation shown in green, and overlap shown in yellow. (**a**, **b**) Low interobserver variability (IV) and low DLS to expert variability (DV) on a scan with no heterogeneities and the both canals were labeled as Clear. (**c**) High IV with a scan labeled with difficult bone structure (DBS) and the left canal was labeled Unclear. (**d**) High DV where the DLS outcome was incomplete with a scan labeled with DBS, movement artifact, and the right canal labeled Unclear.

The qualitative interobserver variance and the variance between the deep learning system and experts can be classified into two categories, i.e. differences in markings and human errors. The differences in markings were caused by the preference of each expert in selecting the middle of the canal and the spacing between markings. This in turn is due to differences in human anatomy such as difficult bone structure, bone porosity, pathological conditions, variable thickness, shape and curvature areas of the canal path, and other sources like metallic structures and imaging artifacts, thus making standardized markings difficult or even impossible. However, this variance turned out not to be significant. In contrast, the variance due to human error was found to be significant, i.e. 29 out of 1200 (i.e. 2.4%) human annotations were outside the actual canal for a few markings, but appeared in the areas not considered to have major clinical relevance. There were no significant differences in errors rate between experts.

# Discussion

To summarize, we have shown that a deep learning system (DLS) can accurately segment the mandibular canal within the interobserver variability for a clinically and technically heterogeneous dataset. In addition, we have demonstrated that the previously introduced system generalizes to three new devices and to a geographically and ethnically separate clinical data.

In the pairwise comparison when measured with the median of the mean curve distance and with the proportion of canal being within 2 mm safety margin, the DLS had similar performance to the pairwise comparison between experts. Specifically, there was better agreement between the DLS and the Expert3 and Expert4, than with the Expert1 and Expert2, which may be due to their role in the development set annotation. We note that the highest agreement was with Expert4 even though most of the development set was annotated by Expert3. When compared with the highest disagreement, the DLS showed lower disagreement than the interobserver disagreement with a statistically significant difference. In the device-wise comparison there were statistically significant differences with lower disagreement for the Promax, DentiScan and GiANO but not for the Scanora3Dx and Viso scanners. When the experts and the DLS were compared to a majority vote consensus segmentation, the DLS gave the lowest median of symmetric mean curve distance with statistically significant difference to the experts. In addition, the deep learning system turned out to have similar variability in the anatomic localization specific error as the human observers have, with the largest errors appearing near the mandibular foramina.

The generalization capability of the artificial intelligence algorithms in radiology is considered to be one of challenges, due to large variances between the imaging parameters, such as protocols, technical solutions, field of view, imaging parameters, and voxel sizes, as well as heterogeneities such as patient anatomy and pathology[2]. In addition, the CBCT image quality is affected by patient movement and metallic artifacts caused by dental or oral surgical materials[16–18]. Despite this, we observed that the deep learning system had similar performance across the different imaging devices as well as a variety of patient specific heterogeneities. When measuring out-of-distribution generalization using the previously published system, which was trained with scans from the ProMax and Scanora 3Dx from a single hospital, the performance was found to be similar across all evaluated devices and all locations for the majority of cases. However, definite conclusions about the deep learning system or interobserver performance in case of ethnic variability are left for future work with larger dataset and controlled study protocol.

The visual quality assessment of the DLS segmentation revealed negligible amounts of errors, which were dissimilar to errors produced by expert annotators. Notable errors produced by the DLS were incomplete canals, whereas the expert errors were mostly deviations of the canal path, in both cases re-annotation would be required. The qualitatively assessed error rate was 1% for the DLS and 2.4% for expert annotators, which demonstrates at least comparable performance to human professionals.

These results indicate that the deep learning system could be utilized in maxillofacial radiology for automatic segmentation of the mandibular canal to augment expert annotators. The applicability of deep learning for mandibular canal segmentation from volumetric imaging data is promising and encourages continuing to limited clinical testing and validation under surveillance of radiologists. In addition, it encourages the application of deep learning approaches for other clinical and research tasks with CBCT and computed tomography scans such as tissue segmentations, cephalometric landmark detection, bone density estimation and utilization of ALARA (As Low AS Reasonably Achievable) principle to minimize harmful effects of the ionizing radiation. For future studies we focus on investigating deep learning system reproducibility and generalization with longitudinal and multimodal patient data.

## Limitations

We acknowledge the following limitation of the deep learning system: Deep learning neural networks that are trained without defining diagnostically important features may have an inherent limitation of possibly learning features that are unknown or ignored by medical experts.

# Conclusions

The deep learning system showed consistently better performance in the localization of the mandibular canal than the radiologists for typical cone beam computed tomography scans. The deep learning system was able to generalize to new imaging devices with a slight lack in robustness for low visibility canals. Further investigation is required in the application of the system for cases with poor image quality and/or canal visibility.

# Data availability

The datasets used in model training, validation, and testing were provided by TAUH and CMU, and as such is not publicly available and restrictions apply to their use according to the Finnish law and General Data Protection Regulation (EU) and to the Thai law, respectively.

# Code availability

The code used for pre- and postprocessing and the deep learning techniques includes proprietary parts and cannot be released publicly. However, the proposed method can be replicated using the information in the Methods section.

# Acknowledgements

This study is based on a retrospective and registration dataset which according to the Finnish law and European General Data Protection Regulation (GDPR) rules, does not need ethical permission. The use of the Finnish clinical imaging data was accepted by the Tampere University Hospital Research Director, Finland. Certificate of Ethical Clearance for the Thai clinical imaging data was given by the Human Experimentation Committee, Faculty of Dentistry, Chiang Mai University, Thailand.

The project was partly supported by Business Finland under project "Digital and Physical Immersion in Radiology and Surgery".

Mandibular canal annotations used in this study were partly provided by Antti Lehtinen, D.D.S. and Mika Mattila, D.D.S., both Specialists of Dentomaxillofacial Dentistry.

# Ethics declarations

## Competing interests

The authors declare no competing interests.

## Author contributions

J.Jä. conception, design of the work, data and annotation acquisition, interpretation of data, data annotation, wrote the main manuscript text; J.S. & J.Ja. conception, design of the work, ran the experiments, analysis, wrote the main manuscript text and produced figures; K.K. conception, interpretation of data, project management, corresponding author, wrote and polished the main manuscript text; H.M. interpretation of data and data annotation; Z.L. analysis, post-processing; A.H. project management, design of the work, analysis; O.S. & V.V. & V.M. project management and design of the work; S.P. & S.N. project management, design of the work, data acquisition; All authors reviewed the manuscript.